\documentclass[runningheads]{llncs}
\usepackage{graphicx}
\usepackage{amsmath}
\usepackage{amssymb}
\usepackage{subfigure}
\usepackage{color}
\usepackage{booktabs}
\usepackage{multirow}
\usepackage{overpic}
\newcommand{\ra}[1]{\renewcommand{\arraystretch}{#1}}
\newcommand{\etal}[1]{et al.}

\begin{document}
\title{A Compact Light Field Camera for Real-Time Depth Estimation}
\author{Yuriy Anisimov \and
Oliver Wasenm\"{u}ller \and
Didier Stricker}
\institute{DFKI - German Research Center for Artificial Intelligence \\
Trippstadter Str. 122, 67663 Kaiserslautern, Germany \\
\email{\{yuriy.anisimov, oliver.wasenmueller, didier.stricker\}@dfki.de}
}
\maketitle              
\begin{abstract}
Depth cameras are utilized in many applications.
Recently light field approaches are increasingly being used for depth computation.
While these approaches demonstrate the technical feasibility, they can not be brought into real-world application, since they have both a high computation time as well as a large design.
Exactly these two drawbacks are overcome in this paper.
For the first time, we present a depth camera based on the light field principle, which provides real-time depth information as well as a compact design.
\end{abstract}

\section{Introduction}
\begin{figure}[t]
\centering
\subfigure[Light Field Camera]{\includegraphics[width=0.3\textwidth]{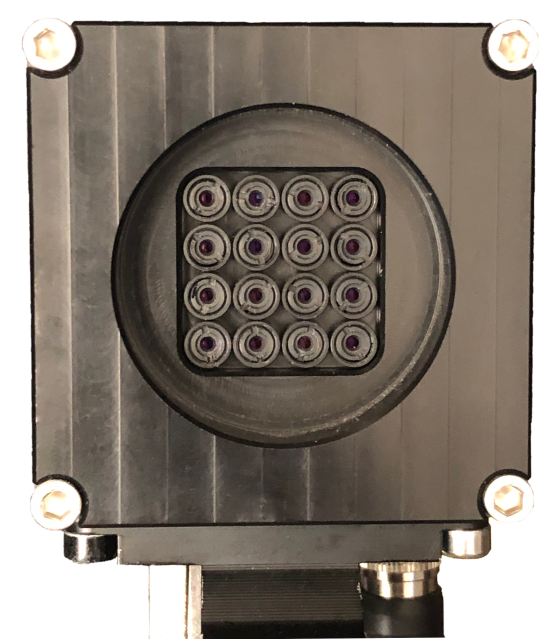}} \hspace{2cm}
\subfigure[Depth Image]{\includegraphics[width=0.4\textwidth]{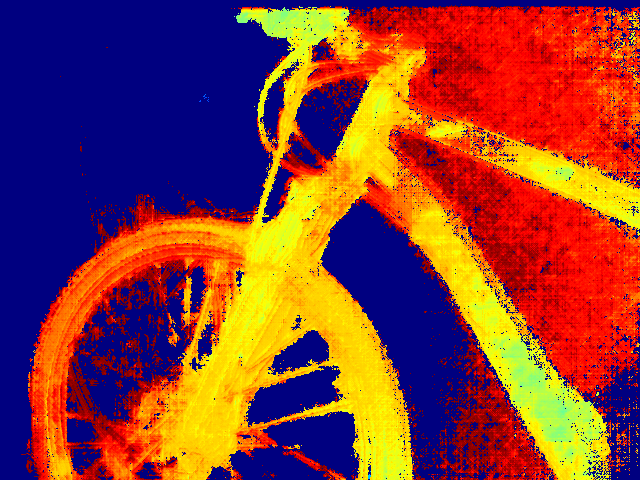}}
\caption{
In this paper, we propose a new compact light field camera, which is capable to compute depth information in real-time.
}
\label{fig:teaser}
\end{figure}

In the recent years depth cameras got popular in different applications, such as automotive \cite{Yoshida2017}, industry \cite{Wasenmueller2016}, gaming \cite{Zhang2012}, etc.
The success factor of these cameras is in particular to record depth information, i.e. the distance of the scene to the camera in each pixel and in real time.
The established technologies can basically be divided into two categories: active and passive depth cameras.
Active cameras have high accuracy in poor lighting conditions and low texture in the scene as they emit active light; 
however, they are often larger in design, have a higher energy consumption and exhibit artifacts through the active lighting.
For this reason, passive cameras are often used in commercial systems.
Usually, these are stereo cameras, which are composed of two cameras determining the depths from the computed pixel disparities.
For the quality of these cameras, the baseline, the distance between the individual cameras, is crucial.
As a result, the baseline dictates the design and size of the camera.
As an alternative to stereo cameras, light field cameras have been explored recently.
They either build using a micro-lens array or consist of several cameras, which are arranged in a matrix configuration (see Figure \ref{fig:teaser}b).
From the redundant pixel information, correspondences and thus depth values can be calculated very efficiently and robustly.
In the state-of-the-art such cameras were already described in detail.
However, these systems suffer two major drawbacks:
the depth estimation algorithms are not working in real-time but sometimes require hours per image \cite{Kim2013}; 
also, the camera systems are usually very large and thus prevent the practical use in filigree setups \cite{Sabater2017,Wilburn2005}.
In this paper, we propose a novel system that handles these two disadvantages.
We build a compact light field camera by placing an array of 4x4 single lenses in front of a full format CMOS sensor.
Furthermore, we enable a real-time depth computation by developing the depth algorithm adequately with dedicated design decisions and suitable for embedded System-on-Chip (SoC).

\section{Related Work}
\label{sec:related}
As mentioned previously, our proposed system is a passive depth camera, which means that no light is actively emitted to determine the depth (compared to Time-of-Flight cameras).
Thus, we focus in this section on passive systems.
The most prominent category of passive depth cameras is stereo cameras.
They are composed of two grayscale or color cameras, which are aligned in parallel with a given distance (baseline).
Depth is estimated by computing the displacement (disparity) of all pixel from one camera to the other.
The parallel alignment of the cameras ensures that corresponding pixels always lie in the same row in the other camera.
Thus, the search for correspondence is considerably simplified.
However, this correspondence search for just two pixels is prone to errors in many scenarios.
In addition, a given baseline is required to ensure appropriate depth resolution.
Thus, the baseline defines the size of the camera.
In this paper, we propose a compact light field camera which takes care of these challenges.
Light field cameras have the advantage to estimate depth not out of just two corresponding pixels, but out of corresponding pixels of all cameras (in our case 16).
This redundancy substantially increases the robustness.
In the last years, several light field cameras were presented.
However, some of them are quite large \cite{Sabater2017,Wilburn2005} preventing their usage in filigree machines.
A famous commercial light field camera are the Lytro \cite{lytro} and Raytrix \cite{raytrix} cameras.
They use micro-lenses with different focus to create the light field.
However, their fragility restricts a usage under industrial or automotive conditions.
In addition, the emphasis of this system is to enable a refocus of images as post-processing.
The systems also offer depth computation, but for single images only and not in real-time.
In contrast, our system is robust for many environmental influences and at the same time provides the depth information as a video stream in real-time.
Focusing on the depth estimation algorithm itself, many approaches were recently presented.
In 2013, Kim et al. in \cite{Kim2013} proposed an accurate method for high-resolution light field depth estimation. It utilizes a concept of epipolar plane image, composed of the row or column slices of the light field images within an angular row or column respectively. In their work, the possible expansion of such light field processing for the 4-dimensional light field was shown. Work of Anisimov and Stricker \cite{Anisimov2017} extends the previously mentioned paper. It proposes the generation of the initial disparity map, based on the limited number of light field images, together with the line fitting cost aggregation in the space of the whole light field with utilization of initial disparity map as the border for the computationally intensive search. 

In the last years, the number of computer vision algorithms based on different types of neural networks is growing. An algorithm, proposed by Shin et al. in \cite{Shin2018} is based on convolutional neural network, which uses not the whole light field, but the sets of images in different angular directions. The result of this algorithm can be considered as precise and the run time is relatively low (several seconds). A recently proposed method from \cite{Peng2018} shows a way of CNN-driven depth estimation, which does not require ground truth for training. Nowadays with the recent GPU architectures the real-time performance of light field algorithms can be achieved. A work of Sabater et al. \cite{Sabater2017} shows the method for the rapid processing of real-world light field videos with wide baseline. It is based on normalized cross-correlation for correspondence matching and involves the pyramidal scheme for disparity estimation. An algorithm from \cite{Qin2017} actively involves the distinct features of GPU architecture for the rapid processing of light field images using image shearing and tensor extraction. One of the accurate algorithms was published by \cite{Zhang2016}, it utilizes the concept of spinning parallelogram operation extraction from the light field images and histogram-based processing on them.
\begin{figure*}[t]
\centering
\includegraphics[width=\textwidth]{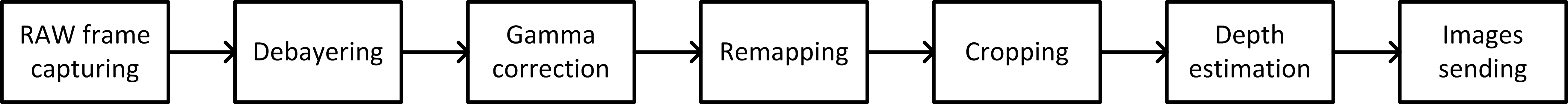}
\caption{
Overview of the single processing steps in our proposed system.
The depth estimation is detailed in Figure \ref{fig:depth}.
}
\label{fig:overview}
\end{figure*}
\section{Hardware}
\label{sec:hardware}
Since we want to design our light field camera as simple as possible, we customize an off-the-shelf full format sensor and equip it with a 4x4 lens array.
In order to create a mobile system for real-time depth estimation, we utilize an embedded GPU-based computational platform.
\subsection{Light Field Camera}
The light field capturing device is built on the base of Ximea CB200CG-CM camera. It uses CMOSIS CMV20000 image sensor with 20 megapixel resolution together with 4x4 single lenses array, placed in the front of the sensor. In this configuration maximum effective resolution of a single light field image is equal to 704x704 pixels. Such a camera assembly approach is easier and more robust to manufacture compared to micro-lens approaches \cite{lytro}. It is somewhat similar to the multi-camera light field capturing devices \cite{Sabater2017}; however, since all of the light field views are captured by a single image sensor, it can be considered as more resilient technology, because cameras synchronization and compensation of single sensors deviations do not need to be performed. The connection of the camera to the computational platform is provided via PCIe interface.

\subsection{Computational Platform}
The Jetson TX2 (see Figure \ref{fig:setup}a) is used as computational platform for our system. It contains the Tegra X2 System-on-Chip, which consists of several ARM cores together with a CUDA-compatible graphics processing unit (GPU). This chip provides support for various peripheral interfaces, such as USB 3.0 or PCIe. Transmission of the processed images from the Jetson board to the recipient is provided by Ethernet interface using the socket-based client-server model. 
%
%
%
%
%
%

\section{Software}
This section describes all software parts of our proposed system. 
The full pipeline is presented in Figure \ref{fig:overview}.  
First, we need to pre-process the images in order to convert the raw information to usable RGB information.
Second, calibration information is utilized to remap all images in order to ensure the epipolar constraint.
As the last step, depth is estimated out of the 16 light field by a sophisticated depth algorithm with dedicated design decisions for real-time performance.
In addition, the depth algorithm is efficiently implemented for the embedded GPU.
\subsection{Pre-Processing}
\label{preprocessing}
The camera used in our system provides images in a RAW sensor format. Thus, some pre-processing steps need to be performed in order to convert images to the proper form for further processing.

The camera sensor captures the light intensity going through the color filter array. In our camera the color filter array utilizes the Bayer filter principle and consists of 2x2 three-colored patterns, which perceive green, blue and red colors. In order to recover the color information from such an image, a so-called debayering needs to be applied. It uses the interpolation of same-colored neighborhood pixels in order to fully reconstruct the intensities values.

After this step, the luminance adjustment, the so-called gamma correction, is performed. A gamma value is selected automatically accordingly to the total scene illumination.
\subsection{Calibration}
\label{calibration}
The camera calibration procedure provides the camera intrinsic values, such as camera focal length and camera center, which are required for the disparity-to-depth conversion, together with the extrinsic values, which represents the camera relative position. Both intrinsics and extrinsics are required for the images rectification, which simplifies the correspondence search problem and hence positively affects the accuracy of the further reconstruction.

The calibration procedure builds upon Zhang's pattern-based algorithm \cite{Zhang2000}, which is extended here to multi-view case. First, the different positions of the calibration pattern are captured by all light field views. The detected coordinates of the pattern are used to estimate the initial single camera calibration values. These values are used for the stereo calibration between a reference view and other light field views. On the estimation of single and stereo calibration the Bundle Adjustment is performed. Based on two views at a maximum distance on the same axis, the reference stereo rectification is calculated. Out of that the global rotation matrix is estimated, and it is used afterwards to align all views to the common image plane. Two remapping tables for all views are generated using the previously computed values. Each of them describes rectified pixels position in X and Y directions respectively.
\begin{figure}[t]
\centering
\includegraphics[width=0.55\textwidth]{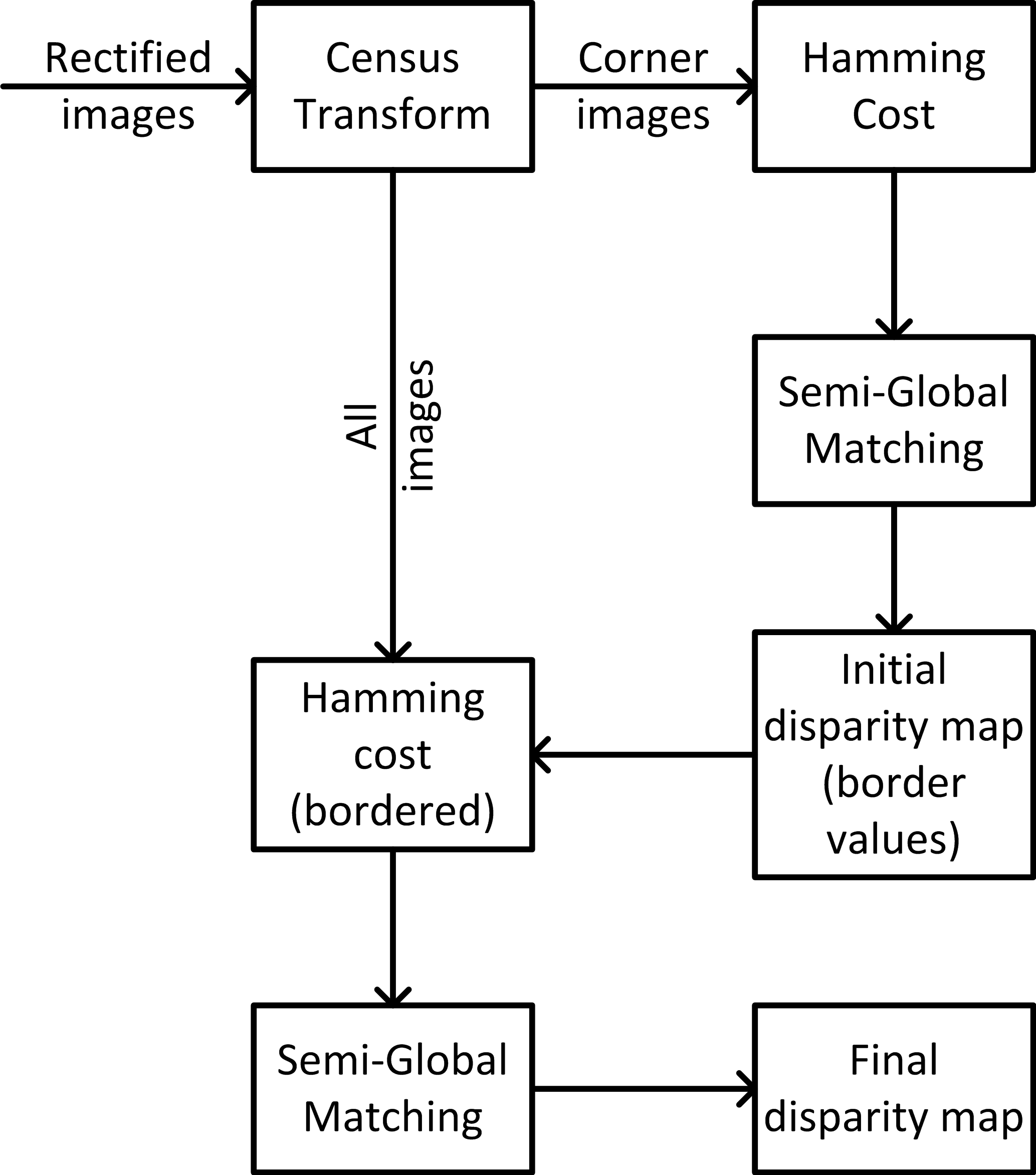}
\caption{
Overview of the proposed real-time depth estimation algorithm.
}
\label{fig:depth}
\end{figure}
\subsection{Efficient Depth Estimation}
\label{depth}
Accurate depth estimation in a short time is a challenging task. Most of the algorithms nowadays are concentrating more on the quality rather than on run time. In this work, we try to concentrate on both aspects. Good depth quality and objects' visual sharpness can be a product of the multi-view principle, while the short run time can be achieved by utilization of modern architectures with a high level of parallelism, such as GPU, together with dedicated design decisions. This depth estimation algorithm utilizes the well-tried computer vision techniques such as coarse-to-fine strategy and the semi-global matching (SGM) method \cite{Hirschmuller2005}. The working flow of the depth estimation algorithm is presented in Figure \ref{fig:depth}.

We consider a two-plane parametrization for the light field, described in \cite{Levoy1996}, by which the light field $L(u,v,s,t)$ is presented by a plane $(s,t)$ for the viewpoints and a plane $(u,v)$ with the pixel information. This representation simplifies the further disparity-related structures estimation. For a given disparity hypothesis $d$ and a reference image $(\hat s, \hat t)$, the pixel position $(u,v)$, corresponding to this disparity in the light-field view $(s,t)$ is defined as
\begin{align}
\hat{p}(u, v, s, t, d) = L(u + (\hat s - s)d, v + (\hat t - t)d, s, t).
\label{align:pix_match}
\end{align}
The algorithm is based on the matching cost, generated from an image similarity measurement. In this work, the Census transform \cite{Zabih1994} is applied to the rectified views. With this operation the radiance value of a pixel in an original view is compared to pixels nearby within a set of pixels coordinates, lying in a window, resulting in a bit string. We use a 3x3 window as a neighborhood.
Taking an image $I$, for every pixel $(u,v)$ in this image, the pixel relation within a neighborhood coordinate set $D$ is performed in a form
\begin{align}
I_{c}(u,v)=\underset{[i,j]\in D}{\bigotimes}\xi(I(u, v),\,I(u+i, v+j)),
\end{align}
where $\otimes$ is the bit-wise concatenation. The pixel relation function $\xi()$ is defined as
\begin{align}
{\xi(p_1, p_2)} =
\begin{cases}
0, &p_1 \leqslant p_2\\
1,& p_1 > p_2
\end{cases}
.
\end{align}
\\
Having a pixel in a reference view, a matching cost function is defined through a Hamming distance between corresponding pixels in a form of two bit strings from Census-transformed images. For two Census-transformed light field images with coordinates $(\hat s, \hat t)$ and $(s, t)$
\begin{align}
C(u, v, d) = HD(L(u, v, \hat s, \hat t), \hat{p}(u, v, s, t, d)),
\label{align:cost_census}
\end{align}
where $HD$ is the Hamming distance function. For two vectors $x_i$ and $x_j$ ($| x_i | = | x_j | = n$, $| \ldots |$ stands for vector cardinality) it can be determined as a number of vectors elements with different values on corresponding positions ($\oplus$ denotes exclusive disjunction).
\begin{align}
HD(x_i, x_j) = \sum\limits_{k=1}^n x_{ik} \oplus x_{jk}.
\label{align:hamming_distance}
\end{align}
Matching cost generation is performed for every pixel in every light field view by
\begin{align}
\begin{split}
S(u,v,d) = {\sum\limits_{s=1}^n\sum\limits_{t=1}^m}{HD(L(u, v, \hat s, \hat t), \hat{p}(u, v, s, t, d))}.
\label{align:S_C}
\end{split}
\end{align}
Out of the matched costs, the final disparity map can be estimated as
\begin{align}
D_s(p) =\underset{d}{\arg\min}\, C_s(p, d)
.
\label{align:wta}
\end{align}
\begin{figure}[t]
\centering
\subfigure[Jetson TX2 Development Kit]{\includegraphics[height=0.23\textheight]{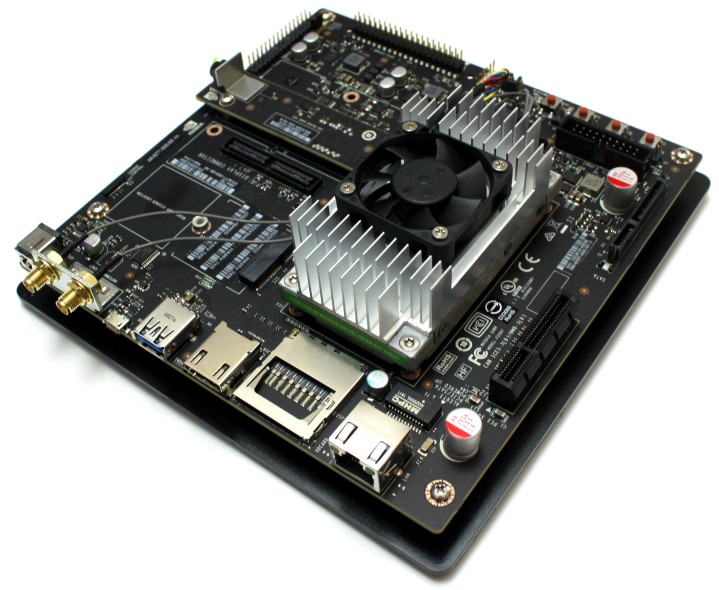}} \hspace{2cm}
\subfigure[Evaluation setup]{\includegraphics[height=0.23\textheight]{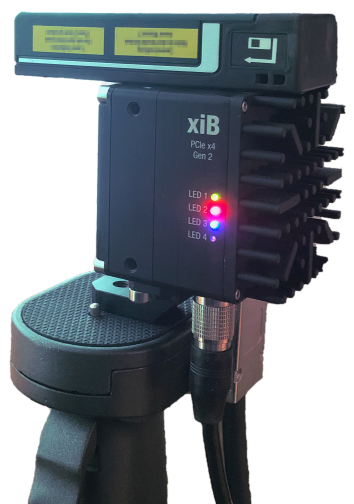}}
\caption{
Our real-time depth estimation is performed on an GPU-based SoC (a). 
For the evaluation of accuracy we added a reference laser scanner (b).
}
\label{fig:setup}
\end{figure}
Performing of such operation can be considered as computationally expensive mainly because of the amount of data involved. One possible strategy could be to reduce the number of processed views; however, that would reduce the disparity map accuracy and object sharpness. Thus, we limit the area of the search for the right hypothesis by obtaining the initial coarse disparity map, based on four cross-lying images in the light field, with respect to reference image $(\hat s, \hat t)$. The results are used for creating the borders for the disparity search range. Having the data for initial disparity $D_{min}$ and $D_{max}$, the range of disparity search in Eq. \ref{align:S_C} is limited to ${d = D_{min}(u,v) - \lambda, ...,D_{max}(u,v) + \lambda}$, where $\lambda$ stands for the bordering threshold, which can be adjusted for changing the degree of impact of the initial disparity map to the final one. With this design decision the execution time is drastically reduced, while the depth accuracy stays at the same high level.
\\
In order to improve quality of the disparity result, to make it smoother and to filter the discontinuities, the SGM method is used for the previously matched cost before applying Eq. \ref{align:wta} to it. The path-wise aggregation for each pixel $p = (u, v)$ and depth hypothesis $d$ in a predefined range, after traversing in direction $r$, formulated as a 2-dimensional vector with the coordinate of a pixel traversing $r$ = \{$\Delta u$, $\Delta v$\}, aggregated cost $L_r$ is  
\begin{align}
\begin{split}
&L_r(p,d) = C(p, d)+\\
&\min\,(L_r(p - r, d),\\
&L_r(p - r, d - 1) + P\mathit{1},\\
&L_r(p - r, d + 1) + P\mathit{1},\\
&\underset{t}{\min}\,L_r(p - r, t) + P\mathit{2}),
\end{split}
\label{align:sgm}
\end{align}
where $P\mathit{1}$ and $P\mathit{2}$ are penalty parameters, $P\mathit{2} \geqslant P\mathit{1}$. 
Traversed costs are then summarized through all traversing directions
\begin{align}
C_s(p,d) = \underset{r}{\sum}\,L_r(p,d).
\label{align:sum}
\end{align} \\
Disparity-to-depth conversion is performed by a classical equation, based on the focal length $f$ and the baseline $b$ between two cameras on one axis at the maximum distance. The principles of light field parametrization allows to do it this way.
\begin{align}
Z(p) =\frac{fb}{D_s(p)}.
\label{align:z}
\end{align}
\\
\begin{figure}[t]
\centering
\includegraphics[height=0.32\textheight]{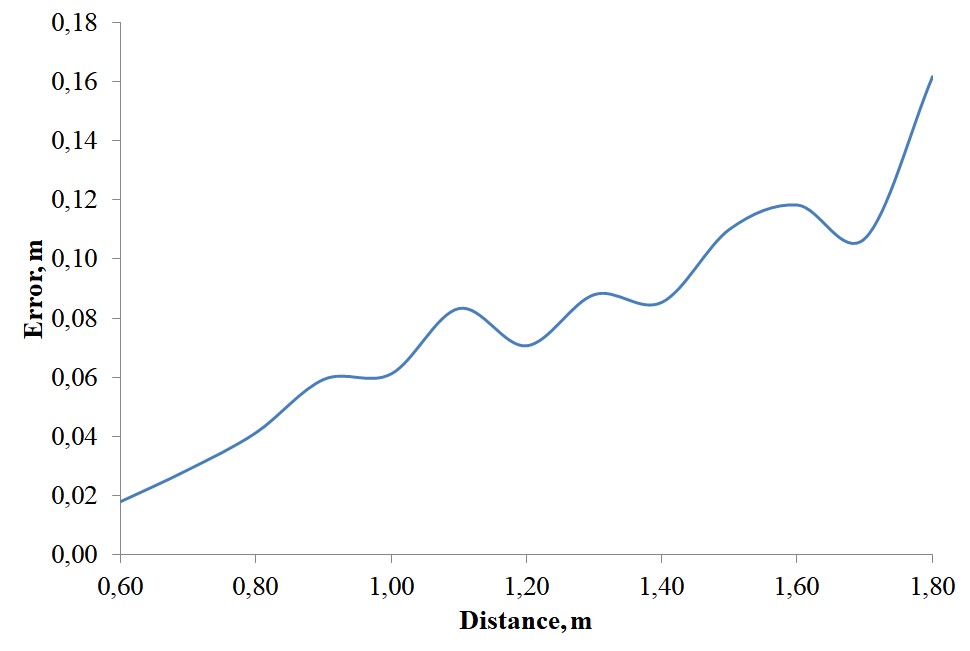}
\caption{
Depth accuracy of our system.
}
\label{fig:acc}
\end{figure}
\subsection{Implementation}
\label{implementation}
The main computations of our system are performed on GPU part of the Tegra TX2. The code is written in C/C++ and uses VisionWorks together with OpenCV libraries. Some of pre-processing steps, such as debayering and gamma correction, are performed by NVIDIA NPP library built-in functions. 
\section{Evaluation}
In this section, we evaluate our novel depth camera system.
We start with a quantitative evaluation and analyze the depth accuracy as well as the run-time of the proposed system.
After that we demonstrate qualitative results of the system.
\begin{table}[t]
\ra{1.3}
\centering
\vspace{2mm}
\caption{
System run time for two different computation platforms. 
Please note that some parts are executed in parallel.
}
\begin{tabular}{lcc} 
\hline
				& Jetson TX2 & GTX 1080 Ti \\
\hline
Capturing		& \multicolumn{2}{c}{30 ms} \\
Preprocessing 	&  45 ms & 16 ms \\
Depth (initial)	&  28 ms & 2 ms \\
Depth (final)	&  100 ms & 8 ms \\
Sending			& \multicolumn{2}{c}{17 ms}\\
\hline
\end{tabular}
\label{tab:runtime}
\end{table}
\subsection{Quantitative Evaluation}
Capturing ground truth data for real-world scenes is a challenging task. 
Using the depth information from other depth cameras can not be considered as a reliable metric, since their result might be flawed as well. 
Thus, we utilize data from an external laser measurement device (Bosch Zamo 3) as shown in Figure \ref{fig:setup}b.
This device has a depth accuracy of $3 mm$, which is one order of magnitude higher than our system accuracy and can thus be considered as reliable.
For the evaluation, we place our camera in front of a flat wall and align the image sensor parallel to it.
In order to avoid any influence of the wall texture, we project twenty different patterns and average the results.
Figure \ref{fig:acc} presents the depth accuracy with respect to the distance. 
The error increases quadratically for higher distances.
For short distances the depth values have an error of below $2 cm$, which is comparable to active devices.
For higher distances the accuracy is still sufficient for many applications.
An even further improved accuracy would be possible by choosing a bigger baseline for the camera matrix.
However, we believe the chosen baseline is the best compromise between camera size and accuracy in the context of the described close range applications.
An evaluation against other light field cameras is not possible, a.o. due to limited availability.
The famous Lytro camera \cite{lytro} is no longer on sale.
In addition, they provide -- like Raytrix \cite{raytrix} -- offline processing of depth data only, which is not comparable with our novel real-time system (cp. Section \ref{sec:related}).
\begin{figure*}[t]
\begin{center}
\centering
\begin{tabular}{ccccc}
\includegraphics[height=27.5mm]{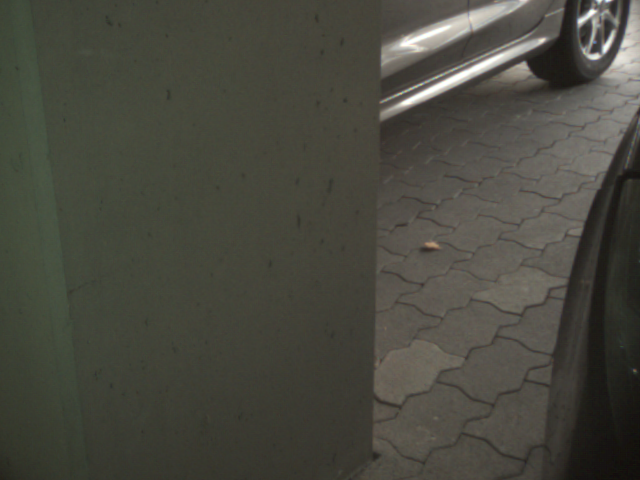} &  
\includegraphics[height=27.5mm]{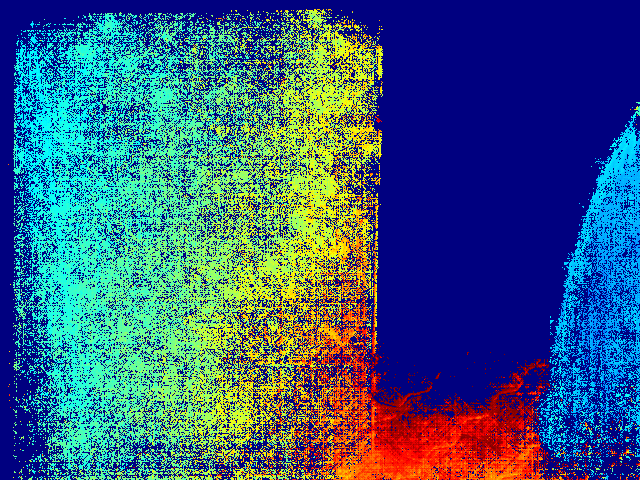} &
\begin{overpic}[height=27.5mm]{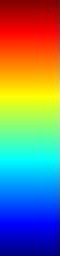} 
\put(0,88){ {\textcolor{white}{far}}}
\put(0,2){{\textcolor{white}{near}}}
\end{overpic}
\\
\includegraphics[height=27.5mm]{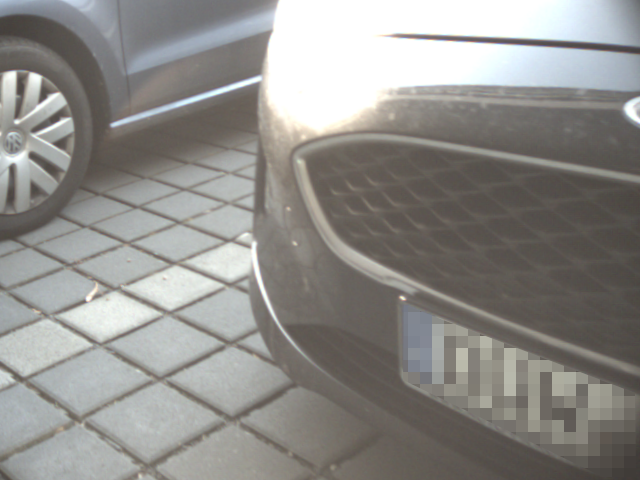} &  
\includegraphics[height=27.5mm]{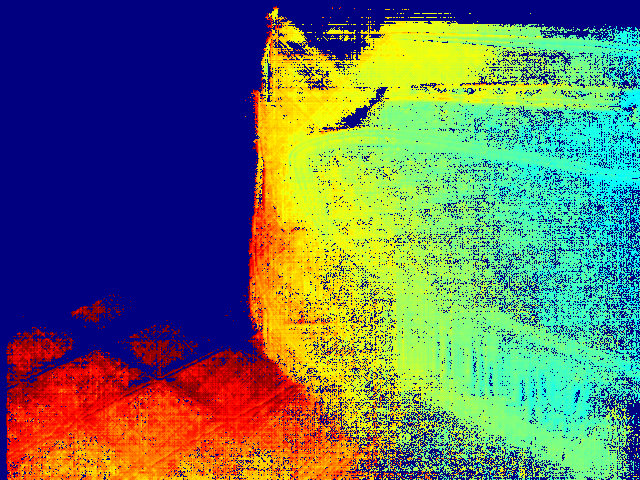}
\\
\includegraphics[height=27.5mm]{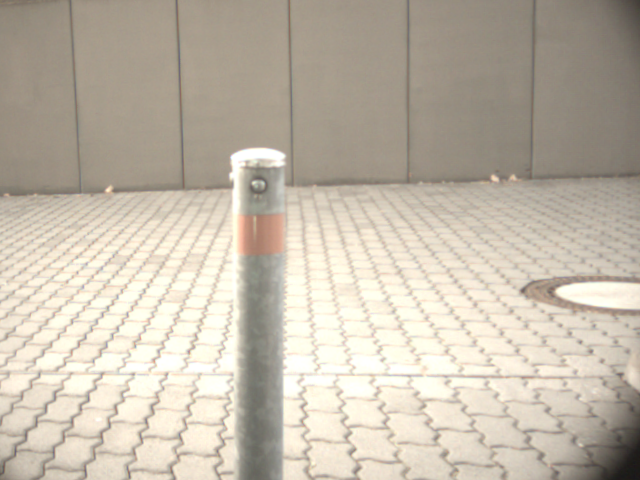} & 
\includegraphics[height=27.5mm]{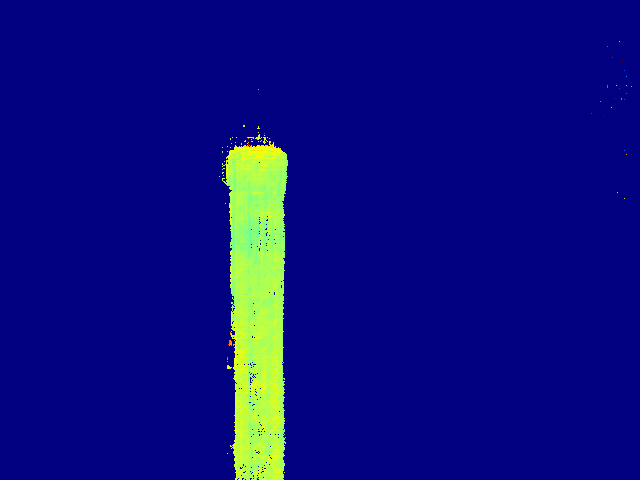}
\\ 
\includegraphics[height=27.5mm]{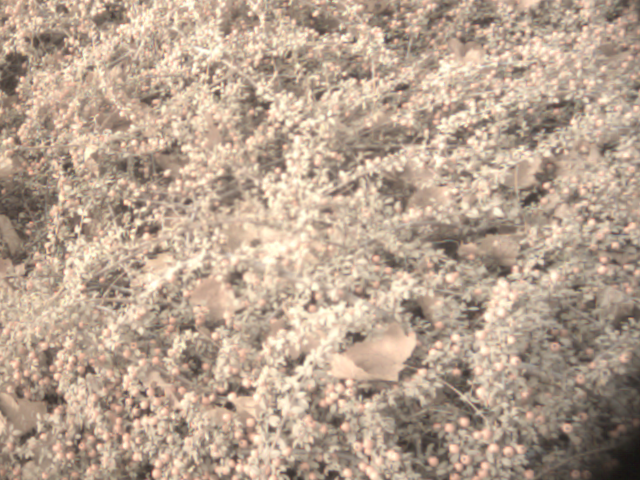} &
\includegraphics[height=27.5mm]{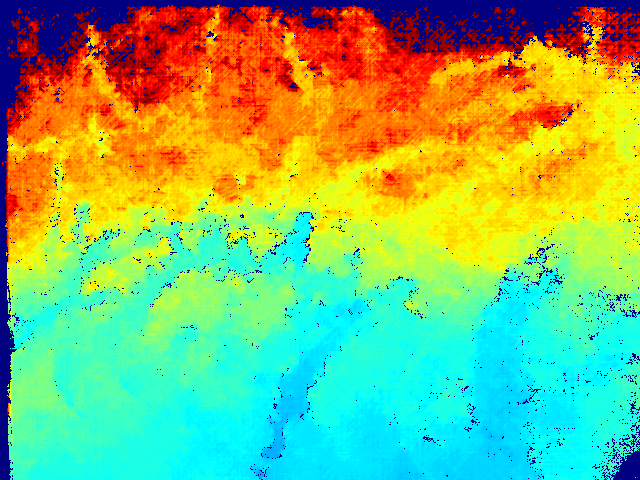}  
\\
\end{tabular}
\end{center}
\caption{
Qualitative results of the proposed system.
The scenes are reconstructed with a high level of detail -- even for homogeneous regions (wall), filigree objects (pillar) and crowded objects (plant hedge).
}
\label{fig:result}
\end{figure*}
\raggedbottom
\subsection{Running Time}
As described in Section \ref{sec:hardware}, our system utilizes a Nvidia Jetson TX2 for embedded processing.
This platform is equipped with a Tegra X2 GPU.
The run times are given in Table \ref{tab:runtime}.
With that setup we achieve a performance of up to $6$ frames per second (FPS), which is sufficient for many applications.
In case higher frame rates are required, a more powerful hardware platform needs to be utilized.
As an example, we evaluate our system running on an ordinary PC (CPU: Intel Xeon E5-1620 v3, GPU: Nvidia GTX 1080 Ti). 
With such a setup up to $32$ FPS can be achieved.
\subsection{Qualitative Evaluation}
Figure \ref{fig:result} shows the qualitative results of our proposed system. 
The depth result is filtered in a way that it keeps distances in our target range of 0.5--2.0 meters only. 
The objects are reconstructed with a high level of detail as visible e.g. for the plant hedge.
In addition, smaller objects like the pillar are robustly detected, which is important for applications such as the automotive domain.
For the flat wall with the homogeneous texture the depth is not very dense, but still depth is sufficiently robust.
\section{Conclusion}
%
%
%
%
%
%
In this paper, we proposed a novel passive depth camera.
Compared to state-of-the-art approaches we provide real-time depth estimation as well as a compact design.
The compact design is realized by using a full format sensor in combination with a lens array in a matrix configuration.
We achieve real-time performance with a novel algorithm utilizing dedicated design decisions and an embedded optimization.
In our evaluation we demonstrate the accuracy of the system in different scenarios.
As the next step, we will work on the density of the depth images.
Some applications require dense depth in order to perform properly \cite{Yoshida2017,Zhang2012}.
Thus, we will have a look at different filtering and interpolation approaches, which were successfully applied in other domains before \cite{Schuster2018}.

\section*{Acknowledgements}
\label{sec:ack}
This work was partially funded by the Federal Ministry of Education and Research (Germany) in the context of the project DAKARA (13N14318).

\end{document}